\documentclass{article}

\usepackage[preprint]{neurips_2022}

\usepackage[utf8]{inputenc} % allow utf-8 input
\usepackage[T1]{fontenc}    % use 8-bit T1 fonts
\usepackage{hyperref}       % hyperlinks
\usepackage{url}            % simple URL typesetting
\usepackage{booktabs}       % professional-quality tables
\usepackage{amsfonts}       % blackboard math symbols
\usepackage{nicefrac}       % compact symbols for 1/2, etc.
\usepackage{microtype}      % microtypography
\usepackage{xcolor}         % colors

\usepackage{multirow}
\usepackage{mathtools}
\usepackage{amssymb}
\setcitestyle{numbers,open={[},close={]}}
\usepackage{placeins}

\title{Deep Neural Networks pruning via the Structured Perspective Regularization}

\author{%
 Matteo Cacciola\\
  CERC, Polytechnique Montréal,\\ Montréal, QC, Canada,\\ 
  \texttt{matteo.cacciola@polymtl.ca} \\
   \And
   Antonio Frangioni \\
   University of Pisa \\
  Pisa, PI, Italy\\
   \texttt{frangio@di.unipi.it} \\
   \AND
   Xinlin Li \\
   Huawei Montreal Research Centre \\
   Montreal, QC, Canada \\
  \texttt{xinlin.li1@huawei.co} \\
  \And
   Andrea Lodi \\
    CERC, Polytechnique Montréal, \\Montréal, QC, Canada,\\ and Jacobs Technion-Cornell Institute, Cornell Tech and Technion - IIT,\\ New York, NY, USA\\
  \texttt{andrea.lodi@cornell.edu} \\
}

\begin{document}

\maketitle

\begin{abstract}
  In Machine Learning, Artificial Neural Networks (ANNs) are a very powerful tool, broadly used in many applications. Often, the selected (deep) architectures include many layers, and therefore a large amount of parameters, which makes training, storage and inference expensive. This motivated a stream of research about compressing the original networks into smaller ones without excessively sacrificing performances. Among the many proposed compression approaches, one of the most popular is \emph{pruning}, whereby entire elements of the ANN (links, nodes, channels, \ldots) and the corresponding weights are deleted. Since the nature of the problem is inherently combinatorial (what elements to prune and what not), we propose a new pruning method based on Operational Research tools. We start from a natural Mixed-Integer-Programming model for the problem, and we use the Perspective Reformulation technique to strengthen its continuous relaxation. Projecting away the indicator variables from this  reformulation yields a new regularization term, which we call the Structured Perspective  Regularization, that leads to structured pruning of the initial architecture. We test our method on some ResNet architectures applied to CIFAR-10, CIFAR-100 and ImageNet datasets, obtaining competitive performances w.r.t.~the state of the art for structured pruning.
\end{abstract}

\section{Introduction}

The striking practical success of Artificial Neural Networks (ANN) has been initially driven by the ability of adding more and more parameters to the models, which has led to vastly increased accuracy. This brute-force approach, however, has numerous drawbacks: besides the ever-present risk of overfitting, massive models are costly to store and run. This clashes with the ever increasing push towards edge computing of ANN, whereby neural models have to be run on low power devices such as smart phones, smart watches, and wireless base stations \cite{NIPS2015_5784, article, DBLP:journals/corr/LengLZJ17}. While one may just resort to smaller models, the fact that a large mode trained even for a few epochs performs better than smaller ones trained for much longer lends credence to the claim \cite{Li_TrainLarge} that the best strategy is to initially train large and over-parameterized models and then shrink them through techniques such as pruning and low-bit quantization.

Loosely speaking, pruning requires finding the best compromise between removing some of the elements of the ANN (weights, channels, filters, layers, blocks, \ldots) and the decrease in accuracy that this could bring \cite{HHGK, NIPS1992_647, GBHPSNS}. Pruning can be performed while training or after training. The advantage of the latter is the ability of using standard training techniques un-modified, which may lead to better peformances. On the other hand, pruning while training automatically adapts the values of the weights to the new architecture, dispensing with the need to re-train the pruned ANN.

A relevant aspect of the process is the choice of the elements to be pruned. Owing to the fact that both ANN training and inference is nowadays mostly GPU-based, pruning an individual weight may yield little to no benefit in case other weights in the same ``computational block'' are retained, as the vector processing nature of GPUs may not be able to exploit un-structured forms of sparsity. Therefore, in order to be effective pruning has to be achieved simultaneously on all the weights of a given element, like a channel or a filter, so that the element can be deleted entirely. The choice of the elements to be pruned therefore depends on the target ANN architecture, an issue that has not been very clearly discussed in the literature so far. This motivates a specific feature of our development whereby we allow to arbitrarily partition the weight vector and measure the sparsity in terms of the number of partitions that are eliminated, as opposed to just the number of weights.

In this work, we develop a novel method to perform structured pruning during training through the introduction of a Structured Perspective Regularization (SPR) term. More specifically, we start from a natural exact Mixed-Integer Programming (MIP) model of the sparsity-aware training problem where we consider, in addition to the loss and $\ell_2$ regularization, also the $\ell_0$ norm of the structured set of weights. A novel application of the Perspective Reformulation technique leads to a tighter continuous relaxation of the original MIP model and ultimately to the definition of the SPR term. Our approach is therefore principled, being grounded on an exact model rather than based on heuristic score functions to decide what entities to prune as prevalent in the literature so far. It is also flexible as it can be adapted to any kind of structured pruning, provided that the prunable entities are known before the training starts, and the final expected amount of pruning is controlled by the hyper-parameter providing the weight of the $\ell_0$ term in the original MIP model. While our approach currently only solves a relaxation of the integer problem, it would clearly be possible to exploit established Operations Research techniques to improve on the quality of the solution, and therefore of the pruning. Yet, the experimental results show that our approach is already competitive with, and often significantly better than, the state of the art. Furthermore, since we perform pruning during training by just changing the regularization term our approach can use standard training techniques and its cost is not significantly higher than the usual training without sparsification.

\section{Related works}

The field of pruning is experiencing a growing interest in the Machine Learning  (ML) community, starting from the seminal work \cite{han} that obtained unexpectedly good results from a trivial magnitude-based approach. The same magnitude-based approach was extended in \cite{lottery} with a re-training phase where the non-pruned weight are re-initialised to their starting  values.

From the structured pruning side, a possibility is adding to the network parameters a scaling factor for each prunable entity, multiplying all the corresponding parameters; then, sparsity is enforced by adding the $\ell_1$ norm of the scaling factors vector. In \cite{differentiableMask} a pruning mask is defined, i.e., a differentiable approximation of a thresholding function that pushes the scaling factors to 0 when they are lower than a fixed threshold, avoiding numerical issues.

Most of the recently-published state-of-the-art works for structured pruning, such as \cite{chip}, \cite{dnr} and \cite{oto}, still make use of very heuristic approaches to compute the \emph{importance} of an element of the ANN (often its $l_2$ norm). This is usually sub-optimal and our approach is seeking improvement on the theoretical justification side.

MIP techniques have been successfully used in the ANN context, but mostly in applications unrelated to pruning like construction of adversarial examples (with fixed weights) \cite{Fischetti}. In \cite{Vielma}, the approach is extended to a larger class of activation functions and stronger formulations are defined. An exception is \cite{Margarida}, where a score function is defined to asses the importance of a neuron and then a MIP is used to minimize the number of neurons that need to be kept at each layer to avoid large accuracy drops. In \cite{Serra} a MIP is used first to derive bounds on the output of each neuron, which is then used in another MIP model of the entire network to find equivalent networks, local approximations and global linear approximations with less neurons of the original network. Since MIPs are $\mathcal{NP}$-hard, these techniques may have difficulties to scale to large ANNs. Indeed, the pruning method developed in \cite{net-trim} rather solves a simpler convex program for each layer to identify prunable entities in such a way that the inputs and  outputs of the layer are still consistent with the original one. This layer-wise approach does not take into account the whole network at once as our own does.

The link between Perspective Reformulation techniques and sparsification has been recently recognised \cite{DoChLi15, AtGo20}, but typically in the context of regression problems that are much simpler than ANNs. In particular, all the above papers count (the equivalent of) each weight individually, and therefore they do not consider structured pruning of sets of related weights as it is required for ANNs. Furthermore, the sparsification approach is applied to input variables selection in settings that typically have orders of magnitude fewer elements to be sparsified than the present one.

\section{Mathematical model}

%\subsection{Setup}

We are given a dataset $X$, an ANN model architecture whose set of parameters $W = \{ \, w_j \,|\, j \in I \, \}$ is partitioned into disjoint subsets $\{ \, E_i \, \}_{i \in N}$, that we call prunable entities, such that $W = \cup_{i \in N} E_i$, and a loss function $L(\cdot)$. If the value of a parameter $w_j$ is zero it could be eliminated from the model (pruned) but, for the reasons discussed above, we are only interested in pruning the entities $E_i$, which is possible only if $w_j = 0$ for all $j \in E_i$. We should consider a three-objective optimization problem where we: i) minimize the loss, ii) minimize some standard regularization term aiming at improving the model's generalization capabilities, and iii) maximize the number of pruned entities $E_i$. As customary in this setting, we approach this by scaling the three objective functions by means of hyperparameters whose optimal values are found by standard grid-search techniques.

\subsection{Method}

Starting from the usual ML framework with $\ell_2$ regularization, we consider the following MIP
\begin{align}
\small
\min \;& \textstyle L(X,W)+\lambda [ \, \alpha \| \, W \, \|_2^2 + (1-\alpha) \sum_{i \in N} y_i \, ]  \label{origObj} \\
&-M y_i\leq w_j \leq M y_i \qquad w_j \in E_i \quad i \in N \label{bigM} \\
&y_i \in \{0,1\} \qquad \qquad\qquad \;\; i \in N \label{yint}
\end{align}
where $\alpha \in [ \, 0 \,,\, 1 \, ]$ and $\lambda >0$ are scalar hyper-parameters while $M$ is an upper bound on the absolute value of the parameters. The binary variable $y_i$ is 0 if the corresponding prunable entity is pruned, 1 if it is not. The standard ``big-M'' constraints \eqref{bigM} ensure that if $y_i = 0$ then $0 \leq w_j \leq 0$ for all parameters in the entity $E_i$, while if $y_i = 1$ the parameters can take any possible useful value (since $M$ is an upper bound). Hence, the term ``$\sum_{i \in N} y_i$'' represents the $\ell_0$ norm of the structured set of weights, i.e.,
\begin{equation*}
 \sum_{i \in N} \left\{\begin{array}{ll}
                    1 & \text{if } \exists j \in E_i \text{ s.t. } w_j \neq 0 \\
                    0 & \text{otherwise.}
                   \end{array}\right.
 \label{L0}
\end{equation*}
%{\vahid maybe you do not need to number this equation, as you do not refer it in the text. You may check for other equations and remove unnecessary numbers. You also forgot to punctuate all equations.}
%
In general, solving \eqref{origObj}--\eqref{yint} directly is not computationally efficient, and therefore a standard strategy is to consider its \emph{continuous relaxation} whereby \eqref{yint} is relaxed to $y_i \in [ \, 0 \,,\, 1 \, ]$. If $L(\cdot)$ is convex, as in our case, this is easily solvable but its solution in both the $y$ and $w$ variables, which is what one could use to perform the pruning, can be rather different from the optimal solution of \eqref{origObj}--\eqref{yint}, therefore leading to inefficient prunings. To improve on that, we consider the Perspective Reformulation \cite{FrGe06a}  of \eqref{origObj}--\eqref{yint}
\begin{equation}
\hspace{-0.25cm}
\begin{array}{l@{\;}l}
 \min &  \displaystyle
          L(X,W)+
         \lambda \Big[ \alpha \hspace{-0.25cm} \sum_{i \in N, j\in E_i} \hspace{-0.12cm}\frac{w_j^2}{y_i}
                        + ( 1 - \alpha ) \sum_{i \in N} y_i \Big] \\
         & \eqref{bigM} \;,\; \eqref{yint}
\end{array}
\label{PR}
\end{equation}
It can be shown that \eqref{PR} has the same integer optimal solution as the original problem, but its continuous relaxation (the Perspective Relaxation) is ``better'' in a well-defined mathematical sense: its optimal objective value is (much) closer to the true optimal value of \eqref{origObj}--\eqref{yint}, which typically implies that its optimal solution is more similar to the true optimal solution. 

\subsection{Eliminating the y variables}

While one can expect that the solution of the relaxation of \eqref{PR} can provide a better guide to the pruning procedure, its form makes it more difficult to apply standard training techniques to solve it. Following the lead of \cite{FGGP11, FrFG16}, we then proceed at simplifying the model by projecting away the $y$ variables. This amounts to computing a closed formula $\tilde{y}(w)$ for the optimal value of the $y$ variables in the continuous relaxation of \eqref{PR} assuming that $w$ are the optimal weights. When $w$ is fixed, the problem decomposes over the subsets, and therefore we only need to consider each fragment
\[
 \textstyle
 f_i(W,y_i) = \lambda \big[ \, \alpha \sum_{j\in E_i}  w_j^2 / y_i + ( 1 - \alpha ) y_i \, \big]
\]
separately. Since $f_i$ is convex in $y_i$ if $y_i > 0$, we just need to find the root of the derivative
\[
 \frac{\partial f_i(W,y_i)}{\partial y_i}=\lambda \left[ -\alpha \hspace{-0.2cm} \sum_{w_j \in E_i} \frac{w_j^2}{y^2_i}+(1-\alpha) \right] = 0
  \;\;,
\]
that is
\[
 y_i=\sqrt{\frac{\alpha \sum_{w_j \in E_i}w^2_j}{1-\alpha}}
\]
(we are only interested in positive $y$), and then project it on the domain. Note that, technically, $f_i(W,y_i)$ is nondifferentiable for $y_i = 0$ but that value is only achieved when $w_j = 0$ for all $j \in E_i$, in which case the choice is obviously optimal. The constraints that defines the domain of $y_i$ can be rewritten as  $y_i \geq | \, w_j \, | / M$ for all $j \in E_i$, together with $y_i \in[0,1]$; putting everything together, we obtain 
\begin{multline*}
 %\textstyle
 \tilde{y}_i(w) = \min \Bigg\{ \max_{j \in E_i} \frac{| \, w_j \, |}{M} \;,\; \sqrt{\alpha \hspace{-0.1cm} \sum_{j \in E_i} \frac{w^2_j}{1-\alpha}} \;,\; 1 \; \Bigg \},
\end{multline*}
where we note that we do not need to enforce positivity since all the quantities are positive. All in all, we can rewrite the continuous relaxation of \eqref{PR} as
\begin{equation}
 \textstyle
 \min \big\{ \, L(X,W) + \lambda \sum_{i=1}^N z_i(w_i;\alpha,M) \, \big\},
 \label{PRz}
\end{equation}
where $w^i = [ \, w_j \, ]_{j \in E_i}$ and, applying some algebraic simplifications, $z_i(w_i;\alpha,M)$ has the closed formula
%\tiny
%\[z_i(W;\alpha,M)=
%\begin{cases}
% \sum_{w_j \in E_i} \sqrt{\frac{1-\alpha}{\alpha }}\frac{w_j^2}{\sqrt{\sum_{w_j \in E_i}w^2_j}}+\sqrt{(1-\alpha)\alpha}\sqrt{ \sum_{w_j \in E_i}w^2_j} & \text{if }   \frac{|w_j|}{M}\leq \sqrt{\frac{\alpha %\sum_{w_j \in E_i}w^2_j}{1-\alpha}}\leq 1\\
% \sum_{w_j \in E_i} \frac{w_j^2M}{A_i}+(1-\alpha)\frac{A_i}{M} &  \text{if }  \sqrt{\frac{\alpha \sum_{w_j \in E_i}w^2_j}{1-\alpha}}\leq   \frac{A_i}{M}\leq 1\\
%\sum_{w_j \in E_i}w_j^2+(1-\alpha) & \text{otherwise} 
%\end{cases}
%\]
%\normalsize 
%where $A_i= \max\{|w_j|: w_j \in E_i\}$. If we call $w^i$ the vector containing  $\{w_j: w_j \in E_i\}$ we have
%\[z_i(W;\alpha,M)=
%\begin{cases}
% \sqrt{\frac{1-\alpha}{\alpha }}||w^i||_2+\sqrt{(1-\alpha)\alpha}||w^i||_2 & \text{if }   \frac{||w^i||_{\infty}}{M}\leq \sqrt{\frac{\alpha }{1-\alpha}}||w^i||_2\leq 1\\
%\frac{M}{||w^i||_{\infty}}||w^i||^2_2+(1-\alpha)\frac{||w^i||_{\infty}}{M} &  \text{if }  \sqrt{\frac{\alpha }{1-\alpha}}||w^i||_2\leq   \frac{||w^i||_{\infty}}{M}\leq 1\\
% ||w^i||^2_2+(1-\alpha) & \text{otherwise} 
%\end{cases}
%\]
\[
\begin{cases}
 \sqrt{\frac{1-\alpha}{\alpha }}(1+\alpha)||w^i||_2 & \text{if }   \frac{||w^i||_{\infty}}{M}\leq \sqrt{\frac{\alpha }{1-\alpha}}||w^i||_2\leq 1\\
\frac{M}{||w^i||_{\infty}}||w^i||^2_2+(1-\alpha)\frac{||w^i||_{\infty}}{M} &  \text{if }  \sqrt{\frac{\alpha }{1-\alpha}}||w^i||_2\leq   \frac{||w^i||_{\infty}}{M}\leq 1\\
 ||w^i||^2_2+(1-\alpha) & \text{otherwise.} 
\end{cases}
\]
We call $z_i(w_i;\alpha,M)$ the \emph{Structured Perspective Regularization} (SPR) w.r.t.~the structure specified by the sets $E_i$. It is easily seen that the SPR behaves like the ordinary $\ell_2$ regularization in parts of the space but it is significantly different in others. Due to being derived from \eqref{PR}, we can expect (with a proper choice of the hyperparameters) the SPR to promote sparsity---in terms of the sets $E_j$---better than the $\ell_2$ norm. Indeed, SPR for $i \in I$ depends on the $\ell_\infty$ norm of $w_i$, which means that it is zero only if all the components of $w_i$ are zero. In other words, while using the, say, ordinary $\ell_1$ would promote sparsity on each weight individually, the SPR can be expected to better promote structured sparsity as required by our application. However, the solution of \eqref{PRz} can be sought for with all of the usual algorithms for training ANNs (SGD, Adam, etc.), and therefore should not, in principle, be more costly then non-sparsity-inducing training.

\subsection{Minor improvements}

Remarkably, the SPR depends on the choice of $M$, which is, in principle, nontrivial. Indeed, all previous attempts of using PR techniques for promoting sparsity \cite{DoChLi15, AtGo20} have been using the ``abstract'' nonlinear form $( 1 - y_i ) w_i = 0$ of \eqref{bigM} (assuming $E_i = \{ \, i \, \}$ as in those treatments). This still yields the same Perspective Reformulation but it is not conducive to projecting away the $y$ variables as required by our approach. While $M$ could in principle be treated as another hyperparameter, in a (deep) ANN, different layers can have rather different optimal upper bounds on the weights; hence, using a single constant $M$ for all the prunable entities is sub-optimal. The ideal choice would be to compute one constant $M_i$ for each entity $E_i$; however, entities in the same layer are often similar to each other, so we decided to compute a different constant for each layer of the network and then use it for all entities belonging to that layer.

Furthermore, all the development so far has assumed that all prunable entities $E_i$ are equally important. However, this may not be true, since different entities can have different number of parameters and therefore impact differently on the overall memory and computational cost. To take this feature into account, we modify our regularization terms as follows:
\[
 \lambda \sum_{i \in N} \frac{u_i}{\sum_{i \in N} u_i} z_i(w_i;\alpha,M),
\]
where $u_i$ is the number of parameters belonging to entity $E_i$.

Finally, we perform a fine-tuning phase. After the ANN has been trained with the SPR, we perform the pruning (eliminating all the entities where $||w^i||_{\infty}$ is smaller than a given tolerance (see below) and then we re-train the pruned network with the standard $\ell_2$ regularization, starting from the value of the weights (for the non-pruned entities) obtained at the end of the previous phase rather than re-initializing them.

\section{Experiments}

We tested our method on the task of filter pruning in Deep Convolutional Neural Networks; that is, the prunable entities are the filters of the convolutional layers. More specifically, the weights in a convolutional layer with $n_{inp}$ input channels, $n_{out}$ output channels and $k \times k$ kernels is a tensor with four dimensions $(n_{inp},n_{out},k,k)$: our prunable entities correspond to the sub-tensors with the second coordinate fixed, and therefore have $n_{inp} \times k \times k$ parameters.

The code used to run the experiments was written starting from the public repository \url{https://github.com/akamaster/pytorch_resnet_cifar10} and \url{https://github.com/pytorch/examples/tree/master/imagenet}.

\subsection{Datasets, architectures and general setup}

For our experiments, we used 3 very popular datasets: CIFAR-10 \cite{cifar}, CIFAR-100 \cite{cifar} and ImageNet \cite{imagenet}. As architectures, we focused on ResNet \cite{resnet} and Vgg \cite{vgg}, in particular, we used ResNet-18, ResNet-20, Resnet50, ResNet-56 and Vgg-16 for CIFAR datasets and ResNet-18 for the ImageNet dataset.

For all the experiments, we used Pytorch (1.7.1 and 1.8.1) with Cuda, the CrossEntropyLoss and the SGD optimizer with 0.9 momentum. 

After the first phase, we considered as pruned all weights with absolute value lower than 1e-4 and we decided to prune all entities with more than 99\% of parameters pruned.

To get the values of $M_i$'s, we used the maximum absolute values of the weights for each layer of a network with the same architecture but trained without our regularization term (for ResNet-20 and ResNet-56 we trained it, for ResNet-18 we used the pretrained version available from torchvision).

More results ant tables are reported in the appendix to show some ablation studies, additional details.

\subsection{Results on CIFAR-10 and CIFAR-100}

These experiments were performed on a single GPU, either a TESLA V100 32GB or NVIDIA Ampere A100 40GB. The model was trained for 300 epochs and then fine tuned for 200 ones.
The dataset was normalized, then we performed data augmentation through random crop and horizontal flip. Mini batches of size 128 (64 for CIFAR-100) were used for training.
The learning rate was initialized to 1e-1 or 1e-2 and then it was divided by 10 at epochs 120, 200, 230, 250, 350, 400 and 450.

\begin{table}
\caption{Results of our algorithm on CIFAR-10 using ResNet-20}
\label{table:1}
\centering
\begin{tabular}{rrrrrr}
\toprule
L-rate & $\lambda$ & $\alpha$ & Acc. & \multicolumn{2}{r}{Pruned pars (\%)} \\
\midrule
0.1 & 1.1 & 0.01 & 85.56 & 231597 & (85.78)\\
0.1 & 1.1 & 1e-4 &86.35  & 227394 & (84.22)\\
0.1 & 1.7 & 0.3 & 87.33  & 217233 & (80.46)\\
0.1 & 0.8 & 0.1 & 88.14  & 206694 & (76.55)\\
0.1 & 1.1 & 0.3 & 89.47  & 199809 & (74.00)\\
0.1 & 0.5 &0.01 & 90.06  & 187767 & (69.54)\\
0.1 & 0.5 & 0.3 & 91.23  & 140535 & (52.05)\\
0.1 & 0.2  & 0.3 & 92.69 & 64188 & (23.77)\\
\midrule
\multicolumn{3}{c}{Original model} & 92.03 & 0 & (0.00)\\
\bottomrule
\end{tabular}
\end{table}

\begin{table}
\caption{Results of our algorithm on CIFAR-100 using ResNet-20. %{\vahid maybe present results on $\alpha \times 100$ to remove 1e-3, 1e-4 etc.}
}
\label{table:c100rs20}
\centering
\begin{tabular}{rrrrrr}
\toprule
L-rate & $\lambda$ & $\alpha$ & Acc. & \multicolumn{2}{r}{Pruned pars (\%)} \\
\midrule
0.01 & 1.3 & 1e-3 & 65.28 & 149184 & (55.25)\\
0.01 & 1.0 & 0.1 & 66.28 & 123984 & (45.92)\\
0.01 & 1.6 & 0.3 & 66.64 & 104256 & (38.61)\\
0.01 & 1.3 & 0.3 & 68.36 & 84960 & (31.47)\\
0.01 & 0.7 & 0.3 & 69.20 & 42480 & (15.73)\\
\midrule
\multicolumn{3}{c}{Original model} & 68.55 & 0 & (0.00)\\
\bottomrule
\end{tabular}
\end{table}

\begin{table}

\caption{Results of our algorithm on CIFAR-10 using ResNet-56}
\label{table:c10rs56}
\centering
\begin{tabular}{rrrrrr}
\toprule
L-rate & $\lambda$ & $\alpha$ & Acc. & \multicolumn{2}{r}{Pruned pars (\%)} \\
\midrule
0.1 & 1.7 & 1e-4 & 90.12 & 756882 & (89.04)\\
0.1 & 1.1 & 0.01 & 91.35 & 720135 & (84.72)\\
0.1 & 0.5 & 0.01 & 92.55 & 633960 & (74.58)\\
0.01 & 1.1 & 1e-4 & 93.13 & 460800 & (54.21)\\
0.01 & 0.8 & 0.1 & 93.98 & 268128 & (31.54)\\
\midrule
\multicolumn{3}{c}{Original model} & 93.35 & 0 & (0.00)\\
\bottomrule
\end{tabular}
\end{table}

As shown in Table \ref{table:1}, training ResNet-20 on CIFAR-10, we were able to prune more than 23\% of the parameters by still increasing the accuracy of the original model, while we could prune almost 70\% of the model by still preserving more than 90\% accuracy.

Using ResNet-20 on the CIFAR-100 dataset, we could again prune more than 15\% of parameters while improving the accuracy of the original model (Table \ref{table:c100rs20}). Due to the fact that CIFAR-100 is more challenging than CIFAR-10, it was not possible to prune very many parameters without losing a significant amount of accuracy: we could still achieve more than 68\% accuracy by pruning a few more than 30\% of the parameters, but accuracy dropped to less than 67\% if pruning more.

Table \ref{table:c10rs56} reports results on training the ResNet-56 architecture on CIFAR-10: once again pruning about 30\% of the parameters improved accuracy and we could keep more than 93\% accuracy while pruning more than a half of the network.

Finally, note that models Resnet-18, ResNet-50, and Vgg-16 have a number of parameters much higher than the ones seen so far. Interestingly, when we employed these models on the Cifar10 dataset, we were able to prune the majority of the parameters (from 81\% to more than 90\%) without really affecting the accuracy of the ANN, sometimes even increasing it. This is shown in Tables \ref{table:c10rs18}, \ref{table:c10rs50} and \ref{table:c10vgg16}.

\subsection{Results on ImageNet}

These experiments were performed on single TITAN V 8GB GPU. The model was trained for 150 epochs and fine tuned for 50 ones. The preprocessing was the same as for the CIFAR datasets. We used mini batches of 256 and 0.1 learning rate that was divided by 10 every 35 epochs. On the ImageNet tests, as usual for datasets with so many classes, we decided to report also the top5 accuracy, that is the percentage of samples where the correct label was on the 5 higher scored classes by the model.

% \begin{table}

% \caption{Results of our algorithm on ImageNet using ResNet-18}
% \label{table:imnetrs18}
% \centering
% \begin{tabular}{rrrrrrr}
% \toprule
% L-rate & $\lambda$ & $\alpha$ &top1 &top5 & \multicolumn{2}{r}{Pruned pars (\%)} \\
% \midrule

% 0.1 & 0.5 & 0.3 & 70.89 & 89.79 & 652644 & (6.11)\\
% 0.1 & 1.3  & 0.3 & 67.26 & 87.71 & 5796625 & (54.33)\\
% 0.1 & 1.3 & 0.01 &  62.27 & 84.43  & 8607247 & (80.67)\\

% \midrule
% \multicolumn{3}{c}{Original model} & 69.76 & 89.08 & 0 & (0.00)\\
% \bottomrule
% \end{tabular}
% \end{table}

\begin{table}[h]
        \scriptsize
        \parbox[t][][t]{.45\linewidth}{
     \caption{Results of our algorithm on CIFAR-10 using ResNet-18}
\label{table:c10rs18}
\centering
        \resizebox{\linewidth}{!}{%
            \begin{tabular}{rrrrrr}
\toprule
L-rate & $\lambda$ & $\alpha$ & Acc. & \multicolumn{2}{r}{Pruned pars (\%)} \\
\midrule
0.1 & 0.8 & 0.3 & 95.24 & 9125688 & (81.40)\\
0.1 & 1.0 & 0.1 & 95.10 & 9712787 & (86.64)\\
0.1 & 1.4 & 0.01 & 94.66 & 10008403 & (89.28)\\
0.1 & 1.8 & 0.01 & 93.76 & 10176969 & (90.78)\\

\midrule
\multicolumn{3}{c}{Original model} & 95.15 & 0 & (0.00)\\
\bottomrule
\end{tabular}
        }}
        \hspace{0.07\linewidth}
        \scriptsize
        \parbox[t][][t]{.45\linewidth}{
        \caption{Results of our algorithm on CIFAR-10 using ResNet-50}
\label{table:c10rs50}
\centering
        \resizebox{\linewidth}{!}{%
            \begin{tabular}{rrrrrr}
\toprule
L-rate & $\lambda$ & $\alpha$ & Acc. & \multicolumn{2}{r}{Pruned pars (\%)} \\
\midrule
0.1 & 1.4 & 0.3 & 94.90 & 20428879 & (86.37)\\
0.1 & 1.8 & 1e-4 & 94.80 & 21041527 & (88.96)\\
0.1 & 2.0 & 0.001 & 94.63 & 21332414 & (90.19)\\
0.1 & 2.2 & 0.01 & 94.30 & 21636397 & (91.48)\\
0.1 & 2.8 & 0.001 & 93.72 & 21964741 & (92.87)\\

\midrule
\multicolumn{3}{c}{Original model} &94.83  & 0 & (0.00)\\
\bottomrule
\end{tabular}
        }}
    \end{table}

    \begin{table}[h]
        \scriptsize
        \parbox[t][][t]{.45\linewidth}{
        \caption{Results of our algorithm on CIFAR-10 using Vgg-16}
\label{table:c10vgg16}
 \centering
        \centering
        \resizebox{\linewidth}{!}{%
          \begin{tabular}{rrrrrr}
\toprule
L-rate & $\lambda$ & $\alpha$ & Acc. & \multicolumn{2}{r}{Pruned pars (\%)} \\
\midrule
0.1 & 1.4 & 0.1 & 93.53 & 13564323 & (92.18)\\
0.1 & 1.8 & 0.1 & 93.23 & 13722777 & (93.25)\\
0.1 & 1.8 & 1e-4 & 93.11 & 13796586 & (93.75)\\
\midrule
\multicolumn{3}{c}{Original model} & 94.12 & 0 & (0.00)\\
\bottomrule
\end{tabular}
        }}
        \hspace{0.07\linewidth}
        \scriptsize
        \parbox[t][][t]{.45\linewidth}{
        \caption{Results of our algorithm on ImageNet using ResNet-18}
 \label{table:imnetrs18}
 \centering
        \resizebox{\linewidth}{!}{%
            \begin{tabular}{rrrrrrr}
\toprule
L-rate & $\lambda$ & $\alpha$ &top1 &top5 & \multicolumn{2}{r}{Pruned pars (\%)} \\
\midrule

0.1 & 0.5 & 0.3 & 70.89 & 89.79 & 652644 & (6.11)\\
0.1 & 1.3  & 0.3 & 67.26 & 87.71 & 5796625 & (54.33)\\
0.1 & 1.3 & 0.01 &  62.27 & 84.43  & 8607247 & (80.67)\\

\midrule
\multicolumn{3}{c}{Original model} & 69.76 & 89.08 & 0 & (0.00)\\
\bottomrule
\end{tabular}
        }}
    \end{table}

Results on ImageNet using ResNet-18 are reported in Table \ref{table:imnetrs18} and show that even in a very large and difficult dataset our method was able to improve the original model results by a consistent margin, reaching almost 71\% accuracy while pruning more than 6\% of the parameters. Pruning more than 50\% of the network caused a drop of 2.5\% in the accuracy, while a way more consistent decrease happened when we pruned about 80\% of the parameters. 
\FloatBarrier
\subsection{Comparison with state-of-the-art methods}

In this section, we compare our results (denoted as SPR) with some of the state-of-the-art algorithms for structured pruning. We report results from \cite{sss} (denoted by SSS),  \cite{espb} (denoted by EPFS), \cite{l2pf} (denoted by L2PF), \cite{pffec} (denoted by PFFEC), \cite{hrank} (denoted as HRANK), \cite{pfc} (denoted as PFC), \cite{chip} (denoted by CHIP), \cite{dnr} (denoted as DNR), \cite{oto} (denoted as OTO)  and \cite{hfp} (denoted by HFP).

Since not all the above papers reported the results for all our metrics (for example, some works only reported the percentage of parameters pruned), in some cases we had to do some conversions that naturally came with some mild approximation. Moreover, in \cite{sss}, only plots were presented, so we had to approximately deduce the data from some points of the figures (figure 2(a) and figure 2(c) of \cite{sss}, we denote the points as P1, P2, etc.). For ImageNet the top5 accuracy is not reported  in \cite{hfp}, so we marked the corresponding field in our table with a ``N/A". Finally, we report results for different settings of each method as they were reported in the original papers; however, it should be remarked that not all of them are filter pruning methods, some rather being general structured pruning ones.

Regarding ResNet-20 on CIFAR-10, our results in Table \ref{table:c10rs20soa} outperform the other methods in most of the cases, meaning that we could reach equal or better accuracy while pruning a larger amount of parameters. Whenever we do not outperform the other methods we have comparable performances and we can have a better accuracy or sparsity  maintaining the other metric close. This is for instance the case of L2PF that achieved 89.9\% accuracy with 73.96\% sparsity, while we achieved a little higher sparsity (74.00\%) losing a little accuracy (89.47\%).

On CIFAR-100 using ResNet-20, the results in Table \ref{table:c100rs20soa} clearly show that we outperform SSS, as we could achieve more than 68.3\% accuracy wile pruning more than 30\% of parameters while SSS could prune only 14.81\% to obtain a little bit more than 67\% accuracy. In Table \ref{table:c10rs56soa}, we can observe a similar situation to ResNet-20 on CIFAR-10 for ResNet-56 on the same dataset. One of the few results we did not outperform was the HPF 93.30 accuracy with 50\% sparsity but we could obtain a little bit more sparsity (54.21\%) with almost the same accuracy (93.13\%).

\begin{table}
\caption{Results of state of the art method on CIFAR-10 using ResNet-20. }
\label{table:c10rs20soa}
\centering
\begin{tabular}{rrrrr}
\toprule
Method & Setting & Acc. & \multicolumn{2}{r}{Pruned pars (\%)} \\
\midrule
\multirow{4}{*}{SSS} &P1 & 90.80 & 120000 & (44.44)\\
&P2 & 91.60  &  40000 & (14.81)\\
&P3 & 92.00 & 10000 & (3.70)\\
&P4 &  92.50 & 0 & (0.00)\\
\midrule
\multirow{4}{*}{EPFS}
&B-0.6 & 91.91  &  70000 & (24.60)\\
&B-0.8 & 91.50 & 100000 & (36.90)\\
&F-0,05  & 90.83 & 130000 & (51.10)\\
&C-0.6-0.05  & 90.98 & 150000 & (56.00)\\
\midrule
L2PF & LW & 89.90 & 199687 & (73.96)\\
\midrule
\multirow{1}{*}{PFC}
&P1 &90.55 &135000 &(50.00) \\
\midrule
\multirow{3}{*}{SPR}
&$\lambda$1.1-$\alpha$0.3 & 89.47 & 199809 & (74.00)\\
&$\lambda$0.5-$\alpha$0.3 & 91.23 & 140535 & (52.05)\\
&$\lambda$0.2-$\alpha$0.3 & 92.69 & 64188 & (23.77)\\
\bottomrule
\end{tabular}
\end{table}

\begin{table}
\caption{Results of state of the art method on CIFAR-100 using ResNet-20. %{\vahid clean the table, maybe divide "Method" column to two columns, one multirow column for method, and single singlerow column for variants. I think the message is you prune with a similar ratio, but have a larger accuracy! Highlight this in table by making bold the digits that you want to argue on.}
}
\label{table:c100rs20soa}
\centering
\begin{tabular}{rrrrr}
\toprule
Method & Setting & Acc. & \multicolumn{2}{r}{Pruned pars (\%)} \\
\midrule
\multirow{4}{*}{SSS}&P1 & 65.50 & 120000 & (44.44)\\
 &P2 & 67.10 & 40000 & (14.81)\\
 & P3 & 68.10 & 10000 & (3.70)\\
 &P4 & 69.20 & 0 & (0.00)\\
\midrule
\multirow{3}{*}{SPR}&$\lambda$1.0-$\alpha$0.1 & 66.28 & 123984 & (45.92)\\
&$\lambda$1.3-$\alpha$0.3 & 68.36 & 84960 & (31.47)\\
&$\lambda$0.7-$\alpha$0.3 & 69.20 & 42480 & (15.73)\\
\bottomrule
\end{tabular}
\end{table}

\begin{table}
\caption{Results of state of the art method on CIFAR-10 using ResNet-56}
\label{table:c10rs56soa}
\centering
\begin{tabular}{rrrrr}
\toprule
Method & Setting & Acc. & \multicolumn{2}{r}{Pruned pars (\%)} \\
\midrule
\multirow{2}{*}{PFFEC}
&A & 93.10 & 80000 & (9.40)\\
&B & 93.06 & 120000 & (13.70)\\
\midrule
\multirow{5}{*}{EPFS}
&B-0.6 & 92.89 & 240000 & (27.70)\\
&B-0.8 & 92.34 & 500000 & (58.60)\\
&F-0.01 & 92.96 & 170000 & (20.00)\\
&F-0.05 & 92.09 & 510000 & (60.10)\\
&C-0.6-0.05 & 92.53 & 570000 & (67.10)\\
\midrule
\multirow{2}{*}{HFP}
&0.5 & 93.30 & 425000 & (50.00)\\
&0.7 & 92.31 & 608430 & (71.58)\\
\midrule
\multirow{3}{*}{HRank}& P1 & 90.72 & 580000 & (68.10)\\
 &P2 & 93.17 & 360000 & (42.40)\\
&P3 & 93.52 & 140000 & (16.80)\\
\midrule
\multirow{1}{*}{PFC}
&P1 &93.05 &425000 &(50.00) \\

\midrule
\multirow{2}{*}{CHIP}
&P1 &92.05 &600000 &(71.80) \\
&P2 &94.16 &360000 &(42.80)\\
\midrule
\multirow{3}{*}{SPR}
&$\lambda$0.5-$\alpha$0.001 & 92.55 & 633960 & (74.58)\\
&$\lambda$1.1-$\alpha$1e-4 & 93.13 & 460800 & (54.21)\\
&$\lambda$0.8-$\alpha$0.1 & 93.98 & 268128 & (31.54)\\
\bottomrule
\end{tabular}
\end{table}

The results reported in Tables \ref{table:c10rs18soa} and \ref{table:c10rs50soa} show that our approach is very competitive with respect to the very recent state-of-the-art methods such as OTO and DNR, sometimes being able improve them significantly. For example, DNR can only prune less than 82\% of ResNet-18 achieving 94.64\% accuracy, while our method reach more than 95\% accuracy pruning more than 86\% of the network.

Similarly, when training Vgg-16 on Cifar-10, our method beats most of the state-of-the-art ones and has always competitive results. For example, CHIP can never prune more than 88\% of the ANN but our algorithm prunes consistently more than 92\% achieving similar or better accuracy (Table \ref{table:c10vgg16soa}).
% \begin{table}
% \caption{Results of state of the art method on CIFAR-10 using ResNet-18}
% \label{table:c10rs18soa}
% \centering
% \begin{tabular}{rrrrr}
% \toprule
% Method & Setting & Acc. & \multicolumn{2}{r}{Pruned pars (\%)} \\
% \midrule
% \multirow{1}{*}{DNR}
% &P1 &94.64 &9233284 &(82.36) \\
% \midrule
% \multirow{3}{*}{SPR}
% $\lambda$-$\alpha$0.8 & 0.3 & 95.24 & 9125688 & (81.4)\\
% $\lambda$-$\alpha$1.0 & 0.1 & 95.10 & 9712787 & (86.64)\\
% \bottomrule
% \end{tabular}
% \end{table}

% \begin{table}
% \caption{Results of state of the art method on CIFAR-10 using ResNet-50}
% \label{table:c10rs50soa}
% \centering
% \begin{tabular}{rrrrr}
% \toprule
% Method & Setting & Acc. & \multicolumn{2}{r}{Pruned pars (\%)} \\
% \midrule
% \multirow{1}{*}{OTO}
% &P1 &94.40 &21570653 &(91.20) \\
% \midrule
% \multirow{3}{*}{SPR}
% &$\lambda$2.0-$\alpha$0.001 & 94.63 & 21332414 & (90.19)\\
% &$\lambda$2.2-$\alpha$0.01 & 94.30 & 21636397 & (91.48)\\
% \bottomrule
% \end{tabular}
% \end{table}

\begin{table}[h]
        \scriptsize
        \parbox[t][][t]{.45\linewidth}{
     \caption{Results of state of the art method on CIFAR-10 using ResNet-18}
\label{table:c10rs18soa}
\centering
        \resizebox{\linewidth}{!}{%
            \begin{tabular}{rrrrr}
\toprule
Method & Setting & Acc. & \multicolumn{2}{r}{Pruned pars (\%)} \\
\midrule
\multirow{1}{*}{DNR}
&P1 &94.64 &9233284 &(82.36) \\
\midrule
\multirow{2}{*}{SPR}
$\lambda$-$\alpha$0.8 & 0.3 & 95.24 & 9125688 & (81.40)\\
$\lambda$-$\alpha$1.0 & 0.1 & 95.10 & 9712787 & (86.64)\\
\bottomrule
\end{tabular}
        }}
        \hspace{0.07\linewidth}
        \scriptsize
        \parbox[t][][t]{.45\linewidth}{
     \caption{Results of state of the art method on CIFAR-10 using ResNet-50}
\label{table:c10rs50soa}
\centering
        \resizebox{\linewidth}{!}{%
            \begin{tabular}{rrrrr}
\toprule
Method & Setting & Acc. & \multicolumn{2}{r}{Pruned pars (\%)} \\
\midrule
\multirow{1}{*}{OTO}
&P1 &94.40 &21570653 &(91.20) \\
\midrule
\multirow{2}{*}{SPR}
&$\lambda$2.0-$\alpha$0.001 & 94.63 & 21332414 & (90.19)\\
&$\lambda$2.2-$\alpha$0.01 & 94.30 & 21636397 & (91.48)\\
\bottomrule
\end{tabular}
        }}
    \end{table}

\begin{table}
\caption{Results of state of the art method on CIFAR-10 using Vgg-16}
\label{table:c10vgg16soa}
\centering
\begin{tabular}{rrrrr}
\toprule
Method & Setting & Acc. & \multicolumn{2}{r}{Pruned pars (\%)} \\
\midrule
\multirow{1}{*}{PFC}
&P1 &93.63 &7357792 &(50.00) \\
\midrule
\multirow{2}{*}{EPSF}
&F-0.005 &94.67 &10305584 &(69.10) \\
&F-0.001 &93.61 &8225584 &(56.70) \\
\midrule
\multirow{1}{*}{PFEEC}
&P1 &93.40 &9315584 &(64.00) \\
\midrule
\multirow{3}{*}{HRANK}
&P1 &93.43 &12205584 &(82.90) \\
&P2 &92.34& 12075584 &(82.10) \\
&P3 &91.23 &12935584 &(92.00) \\
\midrule
\multirow{3}{*}{CHIP}
&P1 &93.86 &11955584 &(81.60) \\
&P2 &93.72 &12215584 &(83.30) \\
&P3 &93.18 &12815584 &(87.30) \\
\midrule
\multirow{1}{*}{DNR}
&P1 &92.00 &1354863818 &(92.07) \\
\midrule
\multirow{1}{*}{OTO}
&P1 &93.30 &1390622688 &(94.50) \\
\midrule
\multirow{3}{*}{SPR}
&$\lambda$1.4-$\alpha$0.1 & 93.53 & 13564323 & (92.18)\\
&$\lambda$1.8-$\alpha$0.1 & 93.23 & 13722777 & (93.25)\\
&$\lambda$1.8-$\alpha$0.0001 & 93.11 & 13796586 & (93.75)\\
\bottomrule
\end{tabular}
\end{table}

\begin{table}
\caption{Results of state-of-the-art method on ImageNet using ResNet-18}
\label{table:imnetrs18soa}
\centering
\begin{tabular}{rrrrrr}
\toprule
Method & Setting &top1 &top5 & \multicolumn{2}{r}{Pruned pars (\%)} \\

\midrule
\multirow{1}{*}{EPFS}
&F-0.05 & 67.81 & 88.37 & 3690000  & (34.60)\\
\midrule
\multirow{2}{*}{HFP}
&0.20 & 69.15 & N/A & 2354869 & (22.07)\\
&0.35 & 68.53 & N/A & 3976709 & (37.27)\\
\midrule
\multirow{1}{*}{SPR}
&$\lambda$1.3-$\alpha$0.3 & 67.26 & 87.71 & 5796625 & (54.33)\\
\bottomrule
\end{tabular}
\end{table}

On ImageNet using ResNet-18, the results in Table \ref{table:imnetrs18soa} show that even if our method does not outperform the other ones, we were able to prune consistently more parameters without a very large accuracy drop. Likely some additional parameter tuning could lead us to even more competitive results.

%\FloatBarrier
\section{Conclusions and future directions}

Based on an exact MIP model for the problem of pruning ANNs, we proposed a new regularization term, based on the projected Perspective Reformulation, designed to promote structured sparsity. The proposed method is able to prune any kind of structures and the amount of pruning is tuned by appropriate hyper-parameters. We tested our method on some classical datasets and architectures and we compared the results with some of the state-of-the-art structured pruning methods. The results have shown that our method is  competitive.

These results are even more promising in view of the fact that further improvements should be possible. Indeed, we are currently solving the continuous relaxation of our proposed exact starting model, albeit a ``tight'' one due to the use of the Perspective Reformulation technique. By a tighter integration with other well-established MIP techniques, further improvements are foreseeable.

\begin{ack}
This work has been supported by the NSERC Alliance grant 544900- 19 in collaboration with Huawei-Canada
\end{ack}
\FloatBarrier
\bibliography{ref}

%%%%%%%%%%%%%%%%%%%%%%%%%%%%%%%%%%%%%%%%%%%%%%%%%%%%%%%%%%%%

%%%%%%%%%%%%%%%%%%%%%%%%%%%%%%%%%%%%%%%%%%%%%%%%%%%%%%%%%%%%

\appendix

\section{Appendix}

\subsection{Time complexity study}
Our method is composed of two steps, in the first one the computation of an additional regularization term is required, while the second one is a fine-tuning phase that is common to many other pruning methods and is just a normal backpropagation. So, we can do a comparison between the cost of an epoch in the first step and in the second one to have a "worst case scenario" of a comparison with other methods that use a regularization term to prune. From Table \ref{table:times}, we can notice that for easier data sets (small input size) the cost of one epoch with our regularization term is roughly the double of a normal one, while for the hardest data set, so in the contest of real applications, this gap decreases and the two costs are almost the same.

\begin{table}

\caption{Average computation times (seconds) for one epoch with and without the SPR term}
\label{table:times}
\centering
\begin{tabular}{lrr}
\toprule
Architecture and data set & time SPR & time without SPR\\
\midrule
ResNet-20 on CIFAR-10 & 13.05 & 6.51 \\
ResNet-56 on CIFAR-10 & 36.58  & 16.99 \\
ResNet-20 on CIFAR-100 & 22.99 & 11.26\\
ResNet-18 on ImageNet & 2,433.14 & 2,401.05\\
\bottomrule
\end{tabular}
\end{table}

\subsection{Detail on grid search and ablation study}
\begin{table}
\caption{Ablation study on hyperparameter (ResNet-56 on CIFAR-10)}
\label{table:ablation}
\centering
\begin{tabular}{rrrr}
\toprule
$\lambda$ & $\alpha$ & Accuracy & Pruned pars\\
\midrule
1.7 & 1e-4 & 90.12 & 89.04 \\
1.4 & 1e-4  & 88.47 & 87.09 \\
1.7 & 0.3 & 90.59 & 85.76\\
0.5 & 0.3 & 94.04 & 6.03\\
\bottomrule
\end{tabular}
\end{table}

As we stated in the first paragraph of Section 3, $\alpha$ and $\lambda$ hyperparameters are found through grid search. We performed a different grid search for each of the data set / architecture pair we used: for ResNet-20 on CIFAR10, we had $\lambda \in [0.2,0.5,0.8,1.1,1.4,1.7]$ and $\alpha \in [1e-4,1e-3,5e-3,1e-2,0.1,0.3]$; for Resnet-56 on CIFAR-10, we had $\lambda \in [0.5,0.8,1.1,1.4,1.7]$; for Resnet-20 on CIFAR-100, we had $\lambda \in [0.1,0.4,0.7,1.0,1.3,1.6]$ and $\alpha \in [1e-3,1e-2,0.1,0.3] $; finally for Resnet-18 on ImageNet, we had $\lambda \in [0.5,0.8,1.0,1.3]$ and $\alpha \in [1e-3,1e-2,0.1,0.3] $. The $\lambda$ parameter is just the usual regularization parameter, the $\alpha$ parameter instead tunes how much of the regularization is computed in a "structured" way. From Table \ref{table:ablation}, we can see that to obtain the maximum level of sparsity $\lambda$ needs to be high, while $\alpha$ needs to be close to 0, while for lower level of sparsity  can be sufficient and sometimes more effective to just increase $\alpha$. Finally, for very low level of pruning, also $\lambda$ needs to be decreased. \begin{table}
\caption{ Accuracy before and after the finetuing phase (ResNet-18 on CIFAR.10) }
\label{table:fine}
\centering
\begin{tabular}{rrrr}
\toprule
$\lambda$ & $\alpha$ & Accuracy before &Accuracy after \\
\midrule
 1.1 & 0.01 & 82.40 & 85.56 \\
 1.7 & 0.3 & 85.28 &87.33  \\
 1.1 & 0.3 & 88.22 & 89.47  \\
 0.5 & 0.3 & 90.62 &91.23  \\
 0.2  & 0.3 & 92.46 &92.69 \\
\midrule
\multicolumn{2}{c}{Original model} & 92.03 & -\\
\bottomrule
\end{tabular}
\end{table}

Finally, we report an observation on the importance of the finetuning phase. From Table \ref{table:fine}, we can see that this step is crucial to obtain higher accuracy when the pruning determined a big drop in this value, while is less important when the accuracy stays high despite the pruning.

\subsection{Observation on the structure of the pruned network}
 From the experiments, we noticed that our algorithm heavily prunes the last layers of the network. This is due to the fact that the gain in the objective function is bigger for these last layers since their filters contain way more parameters than filters belonging to the first layers. When we allow our pruning algorithm to heavily prune the network at the cost of a consistent accuracy drop or when the model is so over-parametrized that even pruning a lot of parameters slightly affect the accuracy, we notice that all final layers are fully pruned. Differently, when we prune less parameters to avoid big accuracy drops the final layers that are not fully pruned tend to be the same for different configurations of the hyperparameters, i.e., likely identifying essential structures of the model. For example, pruning ResNet-18 on the ImageNet data set, the layer with the last residual connection is not pruned in almost all our experiments.

\end{document}